\documentclass[english]{article}
\usepackage[T1]{fontenc}
\usepackage[latin9]{inputenc}
\usepackage{geometry}
\usepackage{color}
\usepackage{babel}
\usepackage{float}
\usepackage{textcomp}
\usepackage{url}
\usepackage{graphicx}
\usepackage{setspace}
\usepackage[unicode=true,pdfusetitle,
 bookmarks=true,bookmarksnumbered=true,bookmarksopen=false,
 breaklinks=false,pdfborder={0 0 0},pdfborderstyle={},backref=section,colorlinks=true]
 {hyperref}
\hypersetup{
 urlcolor=blue}

\makeatletter
\@ifundefined{date}{}{\date{}}
\makeatother

\begin{document}
\title{Style transfer and classification in hebrew news items}
\author{\textbf{Nir Weingarten}\\
Reichman university, Israel\\
\href{http://nir.weingarten@post.idc.ac.il}{nir.weingarten@post.idc.ac.il}}
\maketitle

\section{Introduction}

Hebrew was classified as a Morphological Rich Language (MRL) by Tsarfaty
et al. (2010). As such prefixes and suffixes are appended to words
to change their grammatical meaning. This structure has been shown
by Tsarfaty et al (2019) to results in inherent morphological ambiguity
in the language. For this reason, amongst others, Hebrew NLP algorithms
lag behind other non MRL languages.

Recent developments in NLP, and new research in the field of Hebrew
NLP bring promising prospects. In this work we implement, modify and
examine two different Transformer based architectures for two Hebrew
NLP tasks. We performed style transfer augmented text generation and
classification of news items using a Vanilla Bert-like transformer,
and a fine-tuned version of HeBert (Chirqui and Yahav, 2020). The
learning in both cases was done over data scraped from online\emph{
}news archives.

\section{Related Works}

Both the text generator and the classifier presented in this work
are self-attention based transformers, using the same architecture
described by Vaswani, Shazeer, Parmar, Uszkoreit, Jones, Gomez, Kaiser
and Polosukhin in their paper Attention is All you Need (2017). Transformers
are a neural net architecture designed to utilize \emph{attention}
(and \emph{attention }only)\emph{ }to harvest context information
in sequence transduction tasks. Older Recurrent neural net models
utilized \emph{cross-attention} to infer context in sequence-to-sequence
transduction tasks. Recurrent neural nets (RNN), Long short term memory
neural nets (LSTM) and gated RNNs (GRUs) all generate a sequence of
hidden states - $h_{t}$ - as a function of the input for time $t$
and the previous hidden state $h_{t-1}$. While this allows the neural
net to learn and utilize context for each time $t$, context from
far away positions is diminished as the same hidden state vector is
overloaded with information. To solve this problem a \emph{cross-attention}
mechanism was developed - \emph{cross-attention }uses a weighted average
of the different hidden states for each position, instead of one overloaded
hidden vector. The weights of this weighted average are learned, allowing
the network to decide when to give more 'attention' to different parts
of the sequence, thus never overloading the hidden vector. Transformers
introduced a new type of attention - \emph{self attention} - which
discards recurrence all together in favor a new attention model. Transformers
are built out of \emph{attention heads }where each input $x$ is embedded
to some $k$ dimensional vector and then undergoes three linear layers
(separately) to produce three new vectors - $Q$ (\emph{Query}), $K$
(\emph{Key}) and $V$ (\emph{Value}). We proceed to compute a dot
product of the \emph{Query} vector of each input $x$ and the \emph{Key}
vectors of all other inputs in the sequence. This yields us a new
vector $A$ the length of the input sequence. $A$ is than used as
a weights vector to compute a context vector $M$ in the same fashion
as in \emph{cross-attention}. Thus transformers can easily and quickly
maintain context from any part of the sequence. Stacking many \emph{attention-heads
}yields more robust learning, and the architecture is well suited
for transfer learning. 

A very popular recent adaptation of transformers in NLP is BERT -
Bidirectional Transformers for Language Understanding (Devlin, Chang,
Lee, Toutanova 2018). BERT is used in both the classifier and partly
in the text generator in this work. BERT is an architecture designed
to allow pretraining of strong and robust language models over unlabeled
text bidirectionally (ie conditioning on both left and right context
in all layers). Such a pre trained BERT model can later be fine-tuned
for a specific NLP task and a specific text corpus (of that same language).
Fine tuning a BERT model is extremely quick and results in state of
the art performance . BERT pre-trained networks consists of many 'blocks'
where each block consists of several self-attention heads a residual
connection, layer normalization and feed forward layers.

A recent major development in the field of Hebrew NLP was the introduction
of HeBERT (Chirqui and Yahav, 2020). HeBert is a new and powerful
pre-trained BERT variant for Hebrew shown to outperform all other
Hebrew NLP models on various language tasks such as Emotion detection,
Fill-the-blank, NER, POS and sentiment analysis. Hebrew is considered
a MRL language and as such needs more processing to disambiguate text.
To solve this task the creators of HeBERT have tested various methods
to tokenize Hebrew text in their paper. Different methods were empirically
tested: char tokenization  , different sub-words based tokenization,
morpheme-based and Finally, word based tokenization. They've observed
similar performance between between different sub-words based variants
for unsupervised tasks, where char-based tokenization   outperformed
other methods in the Fill-in-the-blank task. For supervised tasks
a 30K long sub-word tokenizer based on the YAP parser was used and
shown to yield the best results.

YAP ('Yet Another Parser') - a joint morpho-syntactic parsing framework
for processing Modern Hebrew texts (More and Tsarfaty, 2016). YAP
is an extension of Zhang and Clarks's structure-prediction framework
(2011). YAP receives  a complete \emph{Morphological Analysis }(MA)
of some input text in the form of a \emph{lattice} graph structure.
The \emph{lattice} structure contains all possible morphological analysis
of the input text sequence, and each adjacent path in the graph represents
a possible morphological segmentation of the sequence. Each morphological
segmentation is used to build a dependency tree. Thus there are many
dependencies trees (exponential to the number of possible paths).
YAP proceeds to process the MA it two parallel dependent tasks: \emph{Morphological
Disambiguation }and \emph{Dependency parsing:} The task of \emph{Morphological
Disambiguation} (MD) is to produce the most likely lattice-path out
of a \emph{Morphological Analysis} (MA). The task pf \emph{Dependency
parsing} (DEP) is to select the mpst likely dependency tree for a
given path. For some text input $x$ YAP jointly predicts a tuple
$\left(MD(x),DEP(x)\right)$. The predicted MD and DEP correspond
with one another and are the analysis of the sentence with the highest
probability.

Another variant of the original transformer model that was considered
as a base for the classification task is the Multilingual Bert - mBert
(Devlin, 2018). mBert was trained on 104 different languages from
different families and has been shown by Wu and Dredze (2019) to have
high performance as a zero-shot language transfer tasks. HeBert was
shown to outperform mBert in classification tasks as shown in the
HeBert paper.

Uniform manifold approximation and projection (UMAP) Leland, McInnes,
Healy and Melville (2018), is a non-linear dimensionality-reduction
technique for the analysis and presentation of any type of high-dimensional
data. We've used UMAP to visualize embeddings of article titles. UMAP
works by constructing a high dimensional graph representing the data
then optimizes a low-dimensional graph to be as structurally similar
as possible. This problem was previously addressed by T-SNE. The UMAP
algorithm has two stages: 1. Building a weighted graph between nodes
(data points) where weights are the likelihood that two points are
connected (optimizing the weights). This is done with a balance between
global and local considerations. Locally, weights are calculated so
that a point is 'close' to its n-nearest neighbors (local distance
weigh less in sparsely populated areas and more in dense areas). Globally,
it reduces the likelihood of connection as distance grows. 2. Constructing
a low dimension (layout) graph that is as similar as possible to the
weighted, optimized, graph. 

\section{Solution}

\subsection{General approach}

\subsubsection{Data collection and preprocessing}

Data was collected from online news portal archives using a specially
built python web scraper. We've decided to scrape news data rather
than use an existing Hebrew dataset as it featured a large amount
of labeled data, including meta-data that could be used for style
transfer. Such meta data such as tags appended for each article or
the authors name and time of release would be later used for style
transfer and were collected in later versions of the scraper. Furthermore,
the news portals that feed our collective consciousness are an interesting
corpus to study, as their analysis and tracing might reveal insights
about our society and discourse.

\subsubsection{Article title classification}

A pre-trained HeBert sentiment analysis binary classifier was modified
and fined tuned to classify article titles to the news section of
which they came. 

We've also considered the mBert net for this task, but decided to
follow through with HeBert as it was shown to yield superior results.
The only HeBert variant publically available is the binary sentiment
analysis version which is built for a slightly different task, but
since Bert and HeBert in particular is first trained on an independent
corpus and than only fine tuned to a specific task we assumed the
pre trained model behind the sentiment-analysis model will be a strong
Hebrew language model, regardless of the task for which it was fine-tuned. 

\subsubsection{Text generation and style transfer}

As HeBert uses a YAP based tokenizer we we're unable to fine tune
it for text generation, as the generated tokens will not form a readable
language. Moreover, to include style transfer we would need an untrained
network and preferably one that is easily modified. Thus, we've decided
to train our own Bert-like encoder based text generator using char
tokenization as it was shown in the HeBert paper to yield the best
performance in unsupervised Hebrew tasks. Style transfer was achived
by concating meta data about each char to it's embedded vector. Different
ways to concat this data were tested.

\subsection{Design}

\begin{figure}[H]
\begin{centering}
\includegraphics[width=0.7\columnwidth]{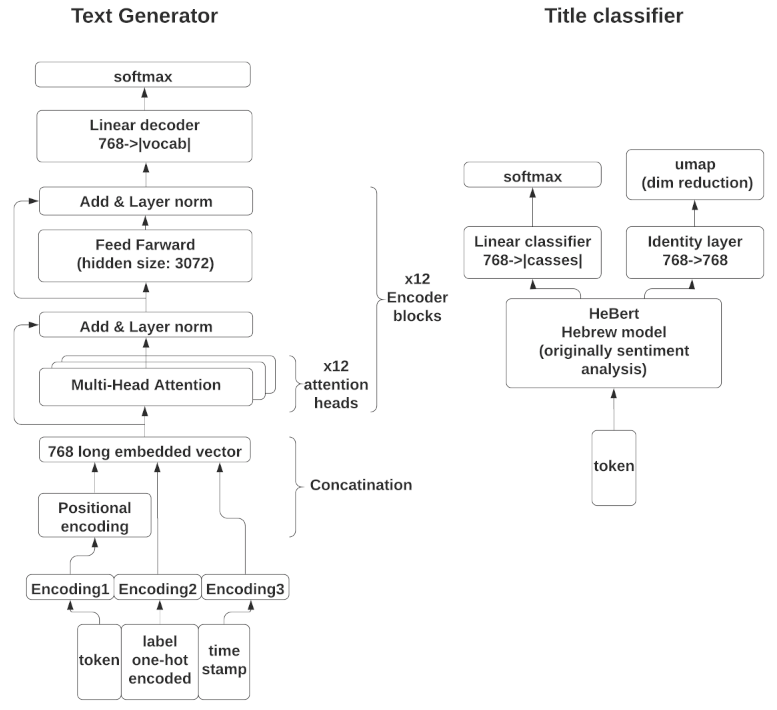}
\par\end{centering}
\caption{Network architecture}
\end{figure}

\subsubsection{Data collection and preprocessing}

We've collected over 1 GB of labeled data from online news portal
archives using a specially built python scraper that iterated over
all articles in the archive since 1.1.2000. We chose to collect articles
from 11 different news sections: (Military, Law and Justice, Health
and Education, World Economy, Israeli Economy, General, Politics,
Soccer, Palestinians, Sex and Sex and relationships) - many other
news sections were  left out. Each article was collected into the
following fields: Main title, Sub title, Article body, Label (news
section), Author, Time of release, Tags added. For text generation
each article was also processed in the following lines: 
\begin{quote}
``{[}SOS{]} main title {[}SEP1{]} sub title {[}SEP2{]} article\_body
{[}PAD{]} {[}EOS{]}'' 

Each lines was truncated to 512 tokens, where the line was shorter
than 512 padding was applied.
\end{quote}

\subsubsection{Article title classification}

To adapt the HeBert sentiment analysis network for a multi-label classification
task we've simply replaced the final linear layer with a new blank
one of the right dimension and used CrossEntropy instead of BinaryCrossEntropy
for a loss function. The original HeBert YAP based 30K long tokenizer
was downloaded from the transformers repository and used to tokenize
the titles. A max length of 50 was chosen for every title. After fine-tuning
the net we proceeded to perform dimensionality reduction on the output
of it's final transformer layer (before the linear classifier). We
will use this reduction to plot a 2D approximation of the representation
of the classified items by our neural net, and gain further insight
about it's inner workings. In title the final 'loaded' token was used
for classification.

\subsubsection{Text generation and style transfer}

The text generator net is built similarly to the Bert model and consists
of an encoding layer, 12 layers of transformer encoder blocks and
a final linear decoder layer activated by softmax. Two different approaches
for embedding were tested: For each token it's 'news section' and
'time of release' data are processed into either a 10d long embedded
vector using two linear layers , or a 2d long vector using simple
min-max normalization. The token itself is processed to either a 758
or 766 long vector corresponding. The token and style vectors were
concatenated to a 768 long vector which was fed into the net. Each
transformer block contains 12 self attention blocks, a layer norm
and residual connection followed by a feed forward network. Finally,
a final soft maxed linear layer was used as a decoder. During training
a square attention mask was used for each 512 long sequence. To generate
text a start string, news section and time stamp are given to the
generator, which recursively feeds the trained model with longer and
longer sequences until the model generates a {[}EOS{]} token or the
512 token limit is reached.

\section{Experimental results}

\subsubsection{Article title classification}

Data was scrambled and split 0.9-0.1 train and validation. The classification
model reached an accuracy of 0.83 on validation set after three training
Epochs using SGD and wAdam optimizer, each epoch taking less than
15 minutes on a Tesla p100-pcie-16gb GPU. Further training did not
improve the results.

Examples of titles that have been correctly classificatied:

\begin{figure}[H]
\begin{centering}
\includegraphics[width=0.5\columnwidth]{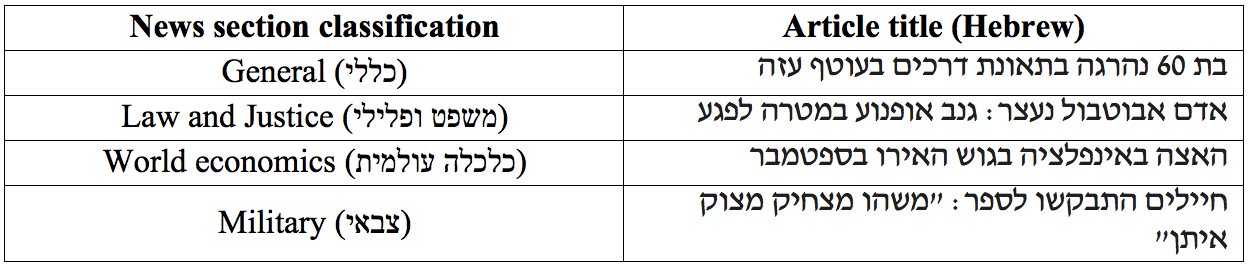}
\par\end{centering}
\caption{Correct classifications}
\end{figure}

Examples of incorrect classifications:

\begin{figure}[H]
\begin{centering}
\includegraphics[width=0.5\columnwidth]{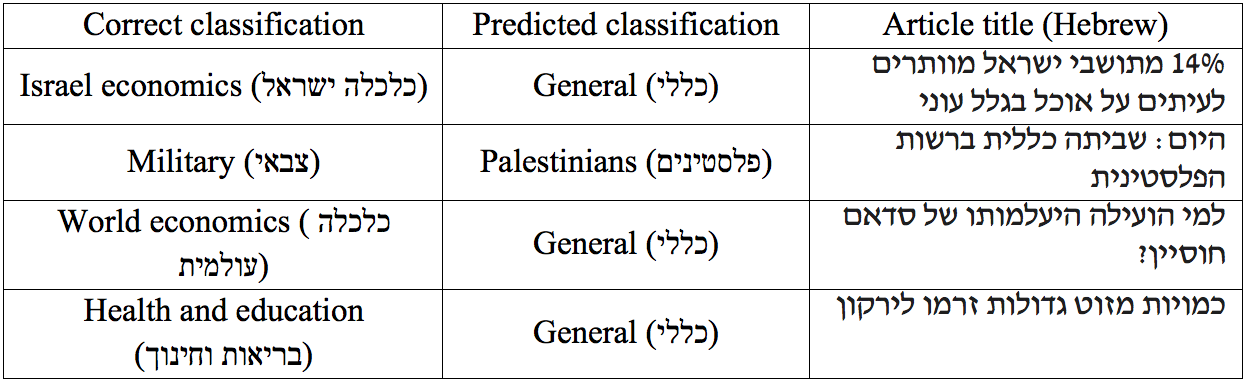}
\par\end{centering}
\caption{Incorrect classifications}
\end{figure}

Casting of each title's latent representation to a 2D array yielded
the following scatter plot:

\begin{figure}[H]
\begin{centering}
\includegraphics[width=0.5\columnwidth]{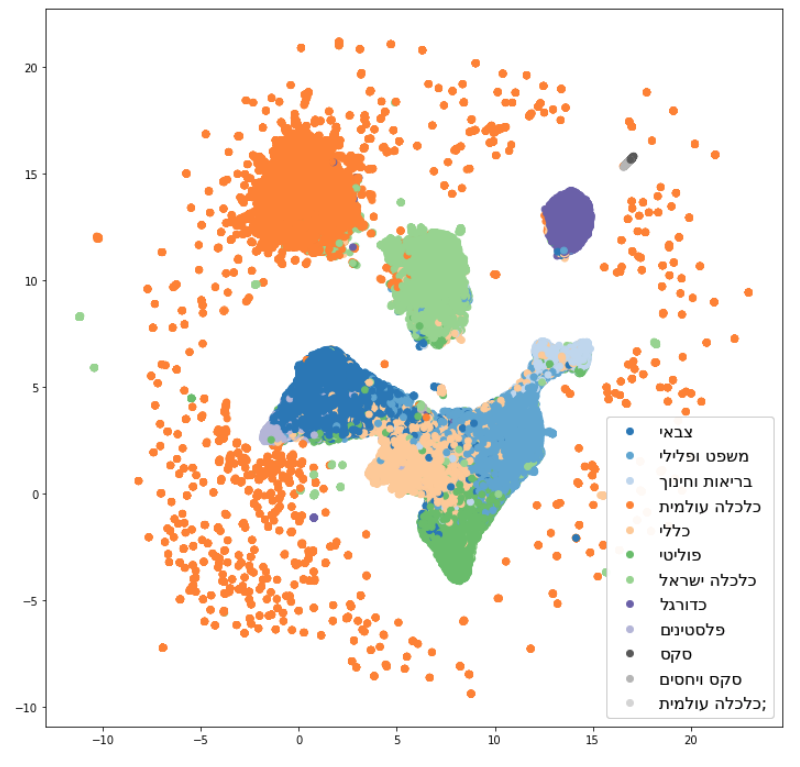}
\par\end{centering}
\caption{Scatter plot of scaled dim reduced latent representations}
\end{figure}

Further analysis of this graph and it's implementations on appendix
A.

\subsection{Article generator}

Both models trained for 20 Epochs on a Tesla p100-pcie-16gb GPU exhibited
very similar loss and preplexity metrics on the validation set. The
model started to overfit after 20 Epochs and was manually stopped
as shown in fig 5. Text generation is an unsupervised task, making
it's performance difficult to measure. We suffice with metrics on
the validation set, qualitative observations and examples. The model
seems to generate coherent sentences   that relate to the starting
vector and in most cases to the news section and time provided to
it. Some outputs have minor grammatical flaws and some tend to recourse
into a loop of the same few word. Most outputs are clear, logical,
exhibit the required style and in some cases seem genuine.
\begin{figure}[H]
\begin{centering}
\includegraphics[scale=0.45]{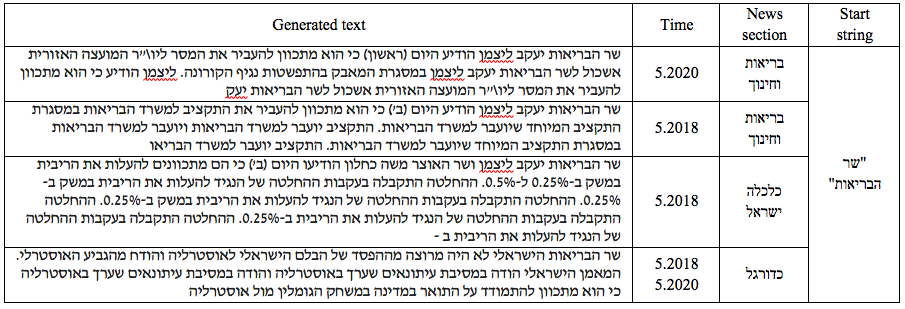}
\par\end{centering}
\caption{Examples of generated texts with style transfer}
\end{figure}

\textbf{}
\begin{figure}[H]
\begin{centering}
\includegraphics[scale=0.5]{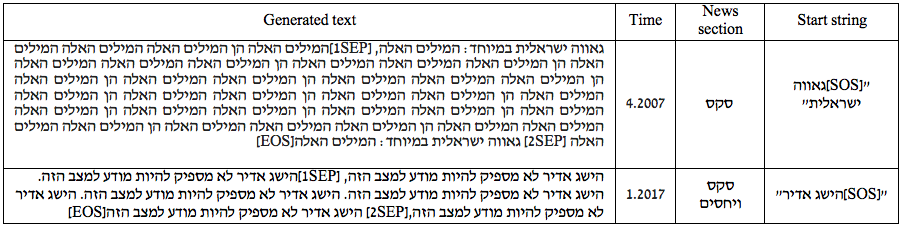}
\par\end{centering}
\textbf{\caption{Poorly generated text}
}
\end{figure}

More examples can be found at \url{www.why-net.net}

\textbf{}
\begin{figure}[H]
\begin{centering}
\includegraphics[width=0.6\columnwidth]{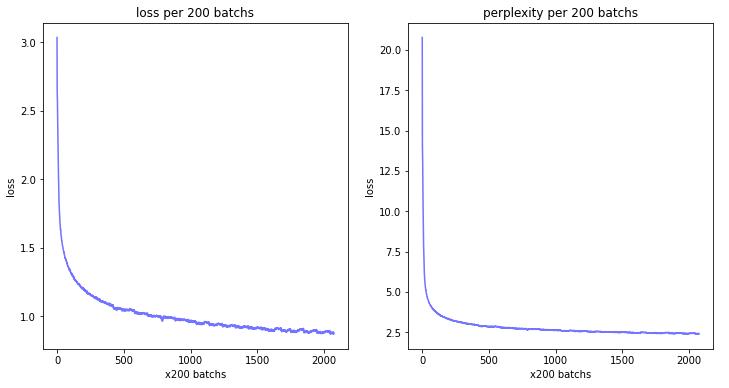}
\par\end{centering}
\textbf{\caption{Train loss and perplexity - text generator}
}
\end{figure}

\section{Discussion}

Despite the ambiguity and morphological richness of Hebrew, it is
evident that transformers in general and Bert in particular are good
tools for quick and transferable Hebrew models. Using the right tokenization
and architecture and sufficient data Transformers can be used to generate
compelling text and accurate classification. Furthermore, we find
that rich style transfer is easily attained in transformers using
simple concatination in the embedding layer. Assuming an accurate
model infers an accurate latent representation of it's items, we can
use the trained model to examine different text sequences in the context
of the learned corpus. Casting such items on a 2D scatter plot yields
a visual tool to hypothesis and deduct qualitative conclusions. For
example, dense clusters can be assumed to be news sections that are
more focused in content, while sparse clusters come from news sections
that cover many varying topics. Close clusters are of news sections
that cover closely related possibly covering the same topics or having
the same writers, far away clusters the opposite. Furthermore, sequences
that are not article titles can be ran through the net and casted
on the scatter plot - assuming all of the above, their location will
reveal their place in the discourse of the corpus that has been learned.
Examples of such castings are shown in appendix A.

Further work can be done on these two models: scraping more data from
more news sections and possibly more news portals, learning style
for author and tags. Char vocabulary should be pruned to allow quicker
learning. Finally, the pretrained generator should be tested as a
pre-trained Hebrew model for various fine tuning NLP tasks. 

\section{Code}

Link to the github repository with the model notebooks:

\url{https://github.com/hopl1t/whynet.git}

\section*{References}

\label{ref:MRL}Tsarfaty R, Seddah D, Goldberg Y, K\textasciidieresis ubler
S, Versley Y, Candito M, Foster J, Rehbein I, Tounsi L (2010) Statistical
parsing of morphologically rich languages (spmrl) what, how and whither.
Proceedings of the NAACL HLT 2010 First Workshop on Statistical Parsing
of Morphologically-Rich Languages, 1--12.

\label{ref:mBert}Jacob Devlin. 2018. Multilingual bert readme document

\label{ref:what is wrong}Reut Tsarfaty, Amit Seker, Shoval Sadde,
Stav Klein. 2019. What is wrong with Hebrew NLP? And how to make it
right

\label{ref:yap}Amir More and Reut Tsarfaty. 2016. Data-driven morphological
analysis and disambiguation for morphologically rich languages and
universal dependencies. In Proceedings of COLING, pages 337--348.
The COLING 2016 Organizing Committee.

\label{ref:structure prediction framework}Yue Zhang and Stephen Clark.
2011. Syntactic processing using the generalized perceptron and beam
search. Computational Linguistics, 37(1):105--151.

\label{ref:hBert}Avihay Chirqui and Inbal Yahav. 2020. HeBERT \&
HebEMO: a Hebrew BERT model and a tool for polarity analysis and emotion
recognition

\label{ref:MD}Yue Zhang and Stephen Clark. 2011. Syntactic processing
using the generalized perceptron and beam search. Computational Linguistics,
37(1):105--151.

\label{ref:DEP}Yue Zhang and Joakim Nivre. 2011. Transition-based
dependency parsing with rich non-local features. In Proceedings of
the ACL, HLT \textquoteright 11, pages 188--193, Stroudsburg, PA,
USA. ACL.

\label{ref:attention is all you need}Vaswani, Shazeer, Parmar, Uszkoreit,
Jones, Gomez, Kaiser, Polosukhin. 2017. Attention is All you Need.

\label{ref:bert}Devlin, Chang, Lee, Toutanova. 2018. BERT: Pre-training
of Deep Bidirectional transformers for Language Understanding.

\label{ref:umap}McInnes, L., Healy, Melville J. 2018. UMAP: uniform
manifold approximation and projection for dimension reduction

\appendix

\section{Appendix A - more scatter plots}

These are scatter plots produced via dimensionality reduction on latent
representations of the titles of articles from the training data.
In each plot an additional word or phrase were ran through the classifier
net and their latent representation cast on the same scatter plot
with the training set titles. The additional plot is in black. We
observe, for example, that sequence ``Benjamin Nethanyahu'' is positioned
in the middle of the Politics section, the sequence ``Gabi Ashkenazi''
is positioned at the border between the Politics and Military sections,
and the sequence ``Moshe Katzav'' is at the border between the Politics
and Law and Justice sections.
\begin{center}
\begin{figure}[H]
\begin{centering}
\includegraphics[width=0.8\paperwidth]{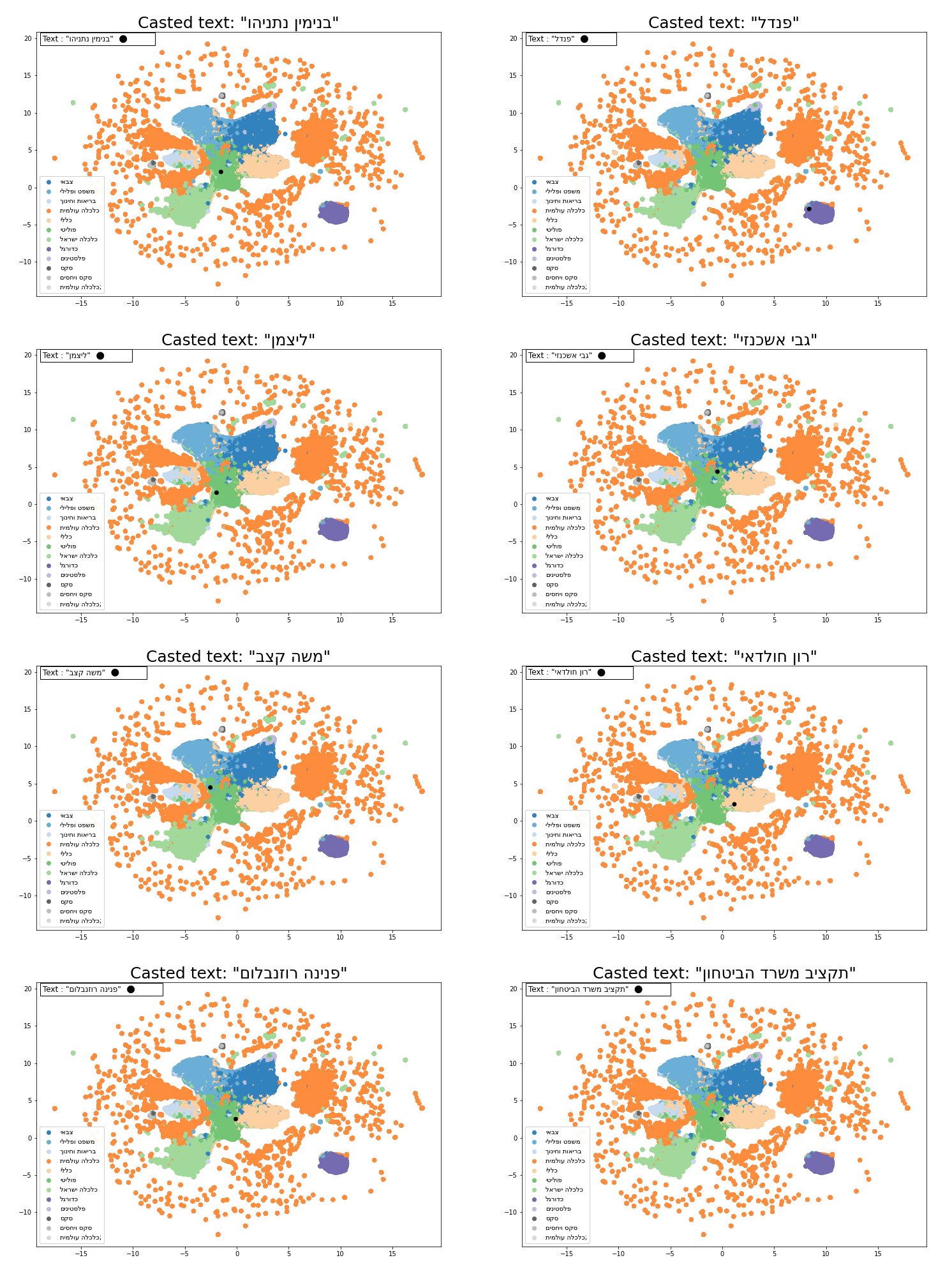}
\par\end{centering}
\centering{}\caption{Casted text}
\end{figure}
\par\end{center}
\end{document}